\documentclass[cameraready]{Interspeech}
\usepackage{amsmath,graphicx}
\usepackage{amssymb}
\usepackage{booktabs}
\usepackage{cite}
\usepackage{microtype}
\usepackage{url}
\usepackage{float}

\allowdisplaybreaks

\title{From Dispersion to Attraction: Spectral Dynamics of Hallucination Across Whisper Model Scales}

\keywords{speech recognition, hallucinations, spectral graph theory, interpretability}

\author[affiliation={1}, orcid=0009-0006-7150-0805, correspondingauthor]{Ivan}{Viakhirev}
\author[affiliation={2,3}, orcid=0009-0001-8203-1059, correspondingauthor]{Kirill}{Borodin}
\author[affiliation={2}, orcid=0000-0002-5802-5513]{Grach}{Mkrtchian}

\address{
    $^1$ Information Technologies, Mechanics and Optics University, Russia\\
    $^2$ lab260, Moscow Technical University of Communications and Informatics, Russia\\
    $^3$ BitmanagerAI, UAE
}
\email{kborodin.research@gmail.com}
\begin{document}

\maketitle

\begin{abstract}
Hallucinations in large ASR models present a critical safety risk. In this work, we propose the Spectral Sensitivity Theorem, which predicts a phase transition in deep networks from a dispersive regime (signal decay) to an attractor regime (rank-1 collapse) governed by layer-wise gain and alignment. We validate this theory by analyzing the eigenspectra of activation graphs in Whisper models (Tiny to Large-v3-Turbo) under adversarial stress. Our results are consistent with the theoretical prediction: intermediate models exhibit Structural Disintegration (Regime I), characterized by a $13.4\%$ collapse in Cross-Attention rank. Conversely, large models enter a Compression-Seeking Attractor state (Regime II), where Self-Attention actively compresses rank ($-2.34\%$) and hardens the spectral slope, decoupling the model from acoustic evidence.
\end{abstract}



\section{Introduction}
Scaling Transformer-based architectures has significantly improved Automatic Speech Recognition (ASR) performance across diverse acoustic conditions~\cite{whisper}. However, scaling introduces systematic failure modes that differ from those in smaller architectures~\cite{Atwany2025DemystifyingHallucination}. While Large Language Models (LLMs) suffer from semantic confabulations~\cite{10.1145/3703155}, ASR models like Whisper~\cite{whisper} exhibit specific transcriptional pathologies triggered by silence, noise, or adversarial stress. These failures range from semantically fluent but acoustically unfounded outputs~\cite{10.1145/3630106.3658996} to severe looping and the generation of memorized training artifacts~\cite{Bara_ski_2025}. Standard performance metrics, such as WER or token log-probabilities, often fail to predict these onsets because the model remains locally confident in its state even when decoupled from the input~\cite{Atwany2025DemystifyingHallucination}.

The problem of hallucinations has been extensively studied through uncertainty estimation and external verification~\cite{10.1145/3571730}. Methods such as semantic entropy~\cite{kuhn2023semantic}, predictive intervals~\cite{lin2022teaching}, and the use of auxiliary LLMs provide tools for post-hoc detection~\cite{li2025tracingrepresentationgeometrylanguage}. However, these approaches are primarily symptomatic, focusing on the statistical properties of the generated text rather than the internal representational dynamics~\cite{10.1145/3571730}. Similarly, theoretical research into Transformer architectures has identified inherent inductive biases toward rank collapse and information ''over-squashing''~\cite{barbero2024transformersneedglassesinformation, jha2025spectralscalinglawslanguage}. While these studies establish the existence of rank degeneration, they typically focus on initialization or general-purpose configurations rather than the specific mechanism of acoustic decoupling in ASR.

In this work, we propose the \textit{Spectral Sensitivity Theorem} to provide a causal explanation for these failures. We demonstrate that hallucination originates from a fundamental phase transition in the geometry of signal propagation: as model depth and structural alignment increase, the mechanism of failure shifts from signal dispersion to attractor dynamics. To analyze this, we introduce the \textit{Spectral Propagation Instability} (SPI) framework, which characterizes how acoustic information is either preserved, scattered, or dominated by internal priors. By validating this theory on Whisper models (Tiny to Large-v3-Turbo) under adversarial stress, we show that while smaller models suffer from structural disintegration (Regime I), larger models enter a compression-seeking attractor state (Regime II), where internal priors actively suppress acoustic evidence.

\section{Theoretical Framework}
\label{sec:theory}

We model the signal propagation through a Transformer network of total depth $L$ as a discrete dynamical system. Let $h_l \in \mathbb{R}^D$ denote the hidden state at layer $l \in \{1, \dots, L\}$. To generalize across architectures (e.g., standard layers vs. Cross-Attention), we formulate the layer function to depend on both the previous state $h_{l-1}$ and an external conditioning context $c$ (such as acoustic features from an encoder). The evolution of the representation and the cumulative Jacobian with respect to the context~\cite{haas2026deepjacobianspectraseparate}, $J_l = \partial h_l / \partial c$, are governed by the recurrences:
\begin{equation}
    h_l = h_{l-1} + f_l(h_{l-1}, c), \quad h_0 = z_0
\end{equation}
\begin{equation}\label{eq:2}
    J_l = W_l J_{l-1} + G_l, \quad J_0 = 0
\end{equation}
where $z_0$ is the initial input embedding, $W_l = I + \partial f_l / \partial h_{l-1}$ is the layer-wise propagator, and $G_l = \partial f_l / \partial c$ represents the layer-wise injection of the external context. We analyze the spectral properties of the final Jacobian $J_L$ to determine the network's sensitivity to acoustic evidence. 

\subsection{Spectral Decomposition and Parameters}
We decompose the propagator $W_j$ into a rank-1 dominant component and a residual perturbation:
\begin{equation}
    W_j = \sigma_{1,j} u_j v_j^T + E_j,
\end{equation}
where $\Vert E_j \Vert_2 = \sigma_{2,j}$. Throughout, $W_j$ denotes the same per-layer propagator $W_l$ from eq.~\ref{eq:2}, re-indexed for the product notation. Three quantities govern how the dominant direction propagates across layers - a relative noise level, an inter-layer alignment, and an effective gain:
\begin{itemize}
    \item \textbf{Spectral gap ($\xi_j$):} $\xi_j = \sigma_{2,j}/\sigma_{1,j} \le \xi \ll 1$. Represents the relative noise level.
    \item \textbf{Alignment ($\kappa_j$):} $\kappa_j = v_{j+1}^T u_j \in [-1, 1]$. Measures geometric coherence between consecutive layers.
    \item \textbf{Effective Gain ($\rho_j$):} $\rho_j = \sigma_{1,j} |\kappa_j|$. Determines signal magnitude stability along the dominant path.
\end{itemize}

\subsection{Lemma 1 (Perturbation Accumulation)}
Let $\Phi^{(M)} = \prod_{j=1}^M W_j$ be the product of $M$ propagators. Define the ideal dominant path as $\Phi_{\text{dom}}^{(M)} = \prod_{j=1}^M (\sigma_{1,j} u_j v_j^T)$. If $\xi_j \le \xi$ and there exists $\varepsilon_\kappa$ such that $|\kappa_j| \ge 1 - \varepsilon_\kappa$ with $\varepsilon_\kappa M \ll 1$, then:
\begin{equation}
    \Phi^{(M)} = \Phi_{\text{dom}}^{(M)} + R^{(M)}, \quad \frac{\Vert R^{(M)}\Vert_2}{\Vert \Phi_{\text{dom}}^{(M)}\Vert_2} \le M\xi + O(M^2\xi\varepsilon_\kappa).
\end{equation}

\noindent\textbf{Proof of Lemma 1.} By induction on $M$. The base case $M=1$ gives $R^{(1)}=E_1$ with relative error $\sigma_{2,1}/\sigma_{1,1}=\xi_1\le\xi$. For the inductive step, expanding $\Phi^{(M+1)}=W_{M+1}\Phi^{(M)}$ and discarding the second-order term $E_{M+1}R^{(M)}$ gives $\Vert R^{(M+1)}\Vert_2 \le \sigma_{1,M+1}\Vert R^{(M)}\Vert_2 + \sigma_{2,M+1}\Vert\Phi_{\text{dom}}^{(M)}\Vert_2$. Dividing by $\Vert\Phi_{\text{dom}}^{(M+1)}\Vert_2 = \sigma_{1,M+1}|\kappa_M|\,\Vert\Phi_{\text{dom}}^{(M)}\Vert_2$ and using $\sigma_{2,M+1}\le\xi\sigma_{1,M+1}$ together with $1/|\kappa_M|\le 1+\varepsilon_\kappa$ yields the recurrence $\varepsilon_{M+1}\le(1+\varepsilon_\kappa)(\varepsilon_M+\xi)$, which unrolls to $\varepsilon_M\le M\xi+O(M^2\xi\varepsilon_\kappa)$ for $\varepsilon_\kappa M\ll1$. \hfill $\square$

\subsection{The Spectral Sensitivity Theorem}
\textbf{Theorem 1.} Let $J_L$ be the accumulated context-sensitivity network Jacobian.

\noindent \textbf{I. Regime I (Disintegration):} If $\rho_j \le \mu < 1$, the sensitivity to early layers decays exponentially, and the Jacobian norm saturates to a bounded constant independent of depth $L$.

\noindent \textbf{II. Regime II (Attractor):} Assume the following conditions hold for depth $L$:
\begin{enumerate}
    \item \textbf{Stability:} $\rho_j \ge 1$ and \textbf{Cumulative Gain:} $\prod_{j=k+1}^L \sigma_{1,j} \ge 1$ for all $k$.
    \item \textbf{Alignment:} $|\kappa_j| \ge 1 - \varepsilon_\kappa$ with $\varepsilon_\kappa L \ll 1$.
    \item \textbf{Injection Coherence:} There exists a direction $\psi_0$ such that for all $k$:
    \begin{itemize}
        \item (a) \textbf{Projection bound:} $\Vert v_{k+1}^T G_k \Vert_2 \ge \gamma \Vert G_k \Vert_F$.
        \item (b) \textbf{Directionality:} $|(v_{k+1}^T G_k)^T \psi_0| / \Vert v_{k+1}^T G_k \Vert_2 \ge 1/\sqrt{2}$.
        \item (c) \textbf{Sign Consistency:} Products $A_{L,k} \cdot (v_{k+1}^T G_k)^T \psi_0$ maintain a constant sign.
    \end{itemize}
    \item \textbf{Spectral Purity:} $\xi L / \gamma \ll 1$.
\end{enumerate}
Then $J_L = u_L \Psi^T + G_L + R_{\Sigma}$, where $\Vert \Psi \Vert_2 = \Omega(\gamma L)$ and the relative noise is bounded by:
\begin{equation}
    \frac{\Vert R_\Sigma \Vert_2}{\Vert \Psi \Vert_2} \le \frac{\xi L}{\gamma}.
\end{equation}

\noindent \textbf{Proof of Theorem 1.}

\noindent \textit{I. Proof of Regime I (Disintegration):} 
By expansion, $J_L = G_L + \sum_{k=1}^{L-1} \Phi_{L,k} G_k$. In this regime, the propagator norm is bounded by the effective gain: $\Vert \Phi_{L,k} \Vert_2 \le \prod_{j=k+1}^L \rho_j$. If $\rho_j \le \mu < 1$, then $\Vert \Phi_{L,k} \Vert_2 \le \mu^{L-k}$. Thus, the contribution of any fixed early injection decays as $\mu^{L-k}$, and the total sensitivity saturates to a bounded constant. The total Jacobian norm is bounded by $\Vert J_L \Vert_2 \le \Vert G \Vert_{\max} \sum_{k=0}^{L-1} \mu^k < \frac{\Vert G \Vert_{\max}}{1-\mu}$. Since the geometric series converges to a constant, the signal does not grow, and long-range dependencies from early acoustic injections are lost.

\noindent \textit{II. Proof of Regime II (Attractor):}
We use Lemma 1 to approximate each path $\Phi_{L,k} \approx A_{L,k} u_L v_{k+1}^T$, where $A_{L,k} = \sigma_{1,L} \prod_{j=k+1}^{L-1} (\sigma_{1,j} \kappa_j)$. The Jacobian expands as $J_L \approx u_L \Psi^T + G_L$, where $\Psi^T = \sum_{k=1}^{L-1} A_{L,k} v_{k+1}^T G_k$.

\noindent\textit{Signal Lower Bound:} Using conditions (3b) and (3c), terms interfere constructively along the coherence direction $\psi_0$:
\begin{equation}
    \Vert \Psi \Vert_2 \ge |\Psi^T \psi_0| = \sum_{k=1}^{L-1} |A_{L,k}| \cdot |(v_{k+1}^T G_k)^T \psi_0|.
\end{equation}
By the projection bound (3a) and $|A_{L,k}| \ge 1$ (from Stability condition):
\begin{equation}
    \Vert \Psi \Vert_2 \ge \sum_{k=1}^{L-1} 1 \cdot \frac{1}{\sqrt{2}} \gamma \Vert G_k \Vert_F \ge \Omega(\gamma L \bar{G}).
\end{equation}

\noindent\textit{Noise Upper Bound:} Residual noise $R_\Sigma = \sum_{k=1}^{L-1} R_{L,k} G_k$ accumulates using the error term from Lemma 1:
\begin{equation}
    \Vert R_\Sigma \Vert_2 \le \sum_{k=1}^{L-1} (L-k)\xi |A_{L,k}| \Vert G_k \Vert_F \approx \xi \frac{L^2}{2} A \bar{G}.
\end{equation}

\noindent\textit{Relative Ratio:} Combining the bounds (assuming approximately constant average amplitude $A$ and injection norm $\bar{G}$):
\begin{equation}
    \frac{\Vert R_\Sigma \Vert_2}{\Vert \Psi \Vert_2} \le \frac{\xi A \bar{G} L^2 / 2}{\gamma A \bar{G} L / \sqrt{2}} = \frac{\xi L}{\gamma} \frac{\sqrt{2}}{2} \le \frac{\xi L}{\gamma}.
\end{equation}
Under the Spectral Purity condition ($\xi L / \gamma \ll 1$), the noise $R_\Sigma$ is negligible relative to the rank-1 component $u_L \Psi^T$, proving the rank-1 collapse. \hfill $\square$

\subsection{Alignment and Geometric Interpretation}

The stability of the Attractor Regime (II) critically depends on the alignment condition $|\kappa_j| \ge 1 - \varepsilon_\kappa$. This requirement is statistically impossible under random initialization. For two independent unit vectors $u, v \in \mathbb{R}^D$ drawn uniformly from the sphere, the expected alignment scales as
\begin{equation}
    \mathbb{E}[|\kappa|] = \mathbb{E}[|u^T v|] \approx \sqrt{\frac{2}{\pi D}}.
\end{equation}
For Whisper Large ($D = 1280$), this yields $\mathbb{E}[|\kappa|] \approx 0.022 \ll 1$. Hence, Regime II cannot arise by chance; it must be created by training dynamics that progressively align dominant subspaces across layers, driving the network toward a structurally coherent, low-rank manifold~\cite{chen2025from}. Our empirical observation of spectral hardening in \texttt{large-v3-turbo} confirms that scaling facilitates this geometric condensation.

Under these conditions, Theorem 1 implies that $J_L$ collapses to a rank-1 structure $u_L \Psi^T$. 
Geometrically, acoustic injections are projected onto a single direction $\Psi$, suppressing orthogonal components and amplifying aligned perturbations.
This produces a structural blindness to acoustic evidence outside the attractor manifold. Consequently, hallucinations in large models are not high-entropy noise but deterministic projections onto a rigid internal direction, effectively decoupling the output from the external context $c$.

\section{Methodology}
\label{sec:method}


To probe the phase transition predicted by the Spectral Sensitivity Theorem, we evaluate the internal representational dynamics of Whisper models under adversarial stress designed to induce acoustic decoupling and attractor lock-in.

\subsection{Adversarial Dataset Generation}
To induce the failure modes described by Regimes I and II, we utilize adversarial perturbations on the LibriSpeech dataset (test-clean and test-other splits). This approach is motivated by recent findings showing that ASR models are highly susceptible to severe hallucination errors under synthetic distribution shifts and noise~\cite{Atwany2025DemystifyingHallucination}. We constructed a ''Hell'' dataset consisting of 5,559 pairs of clean and augmented samples. Following the protocol in~\cite{Bara_ski_2025}, we applied three categories of acoustic stress: 3.5$\times$ time stretch (alignment stress), 6-speaker mixing (manifold saturation), and 0dB Gaussian noise (spectral gap test). We analyze only samples with WER $>$ 0.5 to focus on verified hallucinations.


\subsection{Model Selection and Feature Extraction}
We analyzed three model scales to capture the scaling-induced phase transition: \texttt{tiny} (39M), \texttt{small} (244M), and \texttt{large-v3-turbo} (809M, $D=1280$). For each model, we extracted layer-wise latent representations across depth $L$, focusing on: (i) \textbf{Cross-Attention} to assess acoustic coupling ($G_l$), (ii) \textbf{Self-Attention} to probe contextual priors and attractor formation, and (iii) \textbf{FFN activations} to evaluate latent manifold stability under rank-1 projections.

\subsection{Spectral Observables Predicted by the Theorem}
We use spectral metrics corresponding to Theorem 1. 
For each layer $l$, we compute the singular value decomposition (SVD) of the activation matrix and analyze the spectrum $\Sigma = \{\sigma_1, \dots, \sigma_k\}$ using the following proxies.

     \textbf{Effective Rank ($\mathcal{N}_{\text{eff}}$):} A measure of the effective dimensionality of the representation:
    \begin{equation}
        \mathcal{N}_{\text{eff}} = \exp\left( -\sum \frac{\sigma_i}{\Vert\Sigma \Vert_1} \ln \frac{\sigma_i}{\Vert \Sigma \Vert_1} \right)
    \end{equation}
    A reduction in $\mathcal{N}_{\text{eff}}$ is consistent with the rank-1 collapse ({Regime II}).
    
     \textbf{Spectral Alpha ($\alpha$):} The decay rate of the eigenspectrum. We estimate $\alpha$ via linear regression on log-log coordinates ($\ln \sigma_i \propto -\alpha \ln i$) for the tail of the spectrum. A high $\alpha$ value indicates ''spectral hardening,'' where the lead singular value $\sigma_1$ dominates the tail, validating the condition $\sigma_1 \gg \sigma_2$ (small $\xi$).
    
     \textbf{Kirchhoff Index ($Kf$):} Derived from the spectrum of the activation covariance matrix ($\lambda_i = \sigma_i^2$), representing the effective resistance of the latent graph:
    \begin{equation}
        Kf = \sum_{i=1}^k \frac{1}{\lambda_i}.
    \end{equation}
    An exponential increase in $Kf$ correlates with information ''over-squashing''~\cite{barbero2024transformersneedglassesinformation, jha2025spectralscalinglawslanguage} and signal disintegration as predicted in \textbf{Regime I}.

By tracking these proxies, we can observe the cumulative effect of alignment $\kappa$ and gain $\rho$, providing an empirical link between the mathematical SPI framework and empirical ASR failures.

\section{Results}
\label{sec:results}

\begin{figure*}[t]
  \centering
  \includegraphics[width=\linewidth, trim=0 0 0 2.2cm, clip]{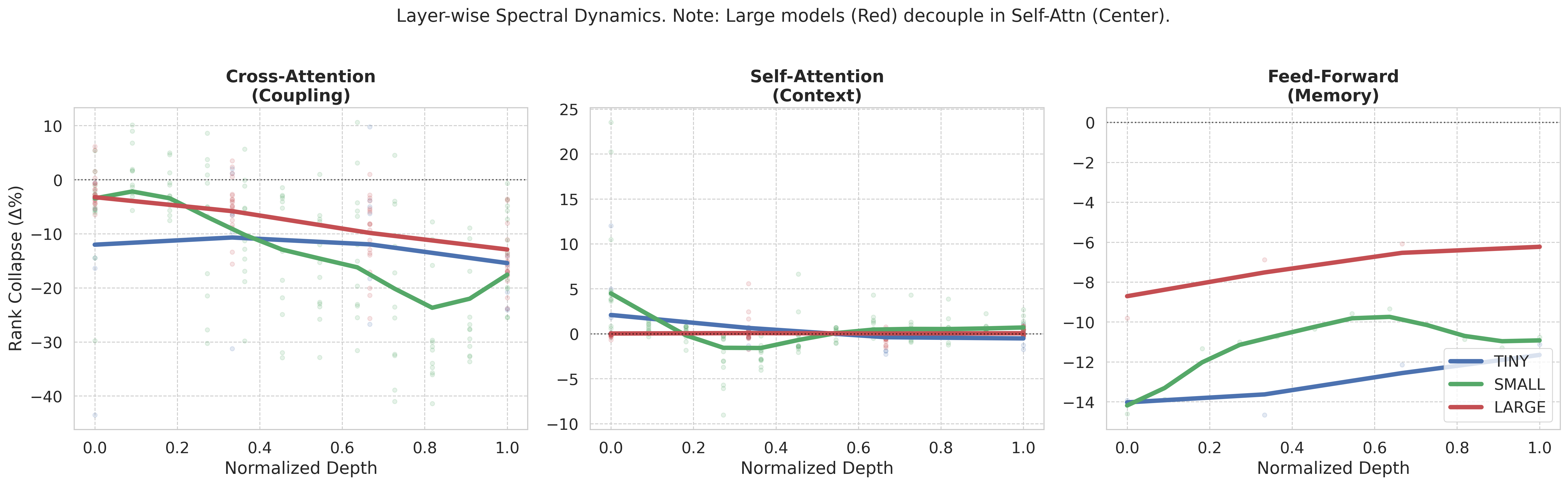}
  \caption{\textbf{Layer-wise Spectral Dynamics.} The evolution of rank collapse ($\Delta \mathcal{N}_{\text{eff}}$) across network depth. \textbf{Left:} Small models (Green) suffer catastrophic decoupling in Cross-Attention (negative slope indicates loss of input signal). \textbf{Center:} Large models (Red) exhibit unique rank compression in Self-Attention, indicative of Attractor dynamics, while Small models expand rank seeking context. \textbf{Right:} FFN rank collapses at every scale, most severely for Tiny and Small, consistent with manifold contraction under perturbation.}
  \label{fig:layer_dynamics}
\end{figure*}

\begin{table}[ht]
  \centering
  \caption{\textbf{Dominant Modes ($K=10$)}. Comparison of spectral shifts between Clean and Adversarial states.}
  \label{tab:topk10}
  \resizebox{\linewidth}{!}{
  \begin{tabular}{l l c c c}
    \toprule
    \textbf{Component} & \textbf{Model} & \textbf{$\Delta \mathcal{N}_{\text{eff}}$ (\%)} & \textbf{$\Delta \log_{10} Kf$} & \textbf{$\Delta \alpha$} \\
    \midrule
    \textbf{Dec Cross-Attn} & Tiny & $-7.56\%$ & $2.12$ & $+0.25$ \\
     & Small & $-9.54\%$ & $2.46$ & $+0.24$ \\
     & \textbf{Large} & $\mathbf{-6.69\%}$ & $2.27$ & $+0.17$ \\
    \midrule
    \textbf{Dec Self-Attn} & Tiny & $-0.76\%$ & $0.49$ & $+0.005$ \\
     & Small & $+0.45\%$ & $3.84$ & $-0.007$ \\
     & Large & $+0.04\%$ & $3.67$ & $-0.007$ \\
    \midrule
    \textbf{Dec FFN} & Tiny & $-0.51\%$ & $2.32$ & $+0.05$ \\
     & Small & $-1.49\%$ & $3.16$ & $+0.08$ \\
     & \textbf{Large} & $\mathbf{-1.78\%}$ & $3.26$ & $+0.10$ \\
    \bottomrule
  \end{tabular}
  }
\end{table}

We analyze the spectral dynamics of hallucination across three model scales. Our findings reveal a distinct dichotomy in failure modes, visualized in Figure~\ref{fig:layer_dynamics} and quantified in Tables~\ref{tab:topk10} and \ref{tab:topk50}. 


\begin{table}[ht]
  \centering
  \caption{\textbf{Spectral Tail ($K=50$)}. The structural breakdown differentiates the scales.}
  \label{tab:topk50}
  \resizebox{\linewidth}{!}{
  \begin{tabular}{l l c c c}
    \toprule
    \textbf{Component} & \textbf{Model} & \textbf{$\Delta \mathcal{N}_{\text{eff}}$ (\%)} & \textbf{$\Delta \log_{10} Kf$} & \textbf{$\Delta \alpha$} \\
    \midrule
    \textbf{Dec Cross-Attn} & Tiny & $-11.98\%$ & $2.93$ & $+0.17$ \\
     & \textbf{Small} & $\mathbf{-13.40\%}$ & $3.87$ & $+0.21$ \\
     & Large & $-4.70\%$ & $3.83$ & $+0.14$ \\
    \midrule
    \textbf{Dec Self-Attn} & Tiny & $-0.10\%$ & $3.38$ & $-0.024$ \\
     & Small & $+0.37\%$ & $4.37$ & $+0.013$ \\
     & \textbf{Large} & $\mathbf{-2.34\%}$ & $4.05$ & $\mathbf{+0.008}$ \\
    \midrule
    \textbf{Dec FFN} & Tiny & $-12.00\%$ & $3.79$ & $+0.16$ \\
     & \textbf{Small} & $\mathbf{-13.59\%}$ & $\mathbf{4.65}$ & $+0.22$ \\
     & Large & $-9.92\%$ & $4.22$ & $+0.21$ \\
    \bottomrule
  \end{tabular}
  }
\end{table}

\subsection{Visualizing the Phase Transition}
\begin{figure}[t]
  \centering
  \includegraphics[width=\linewidth]{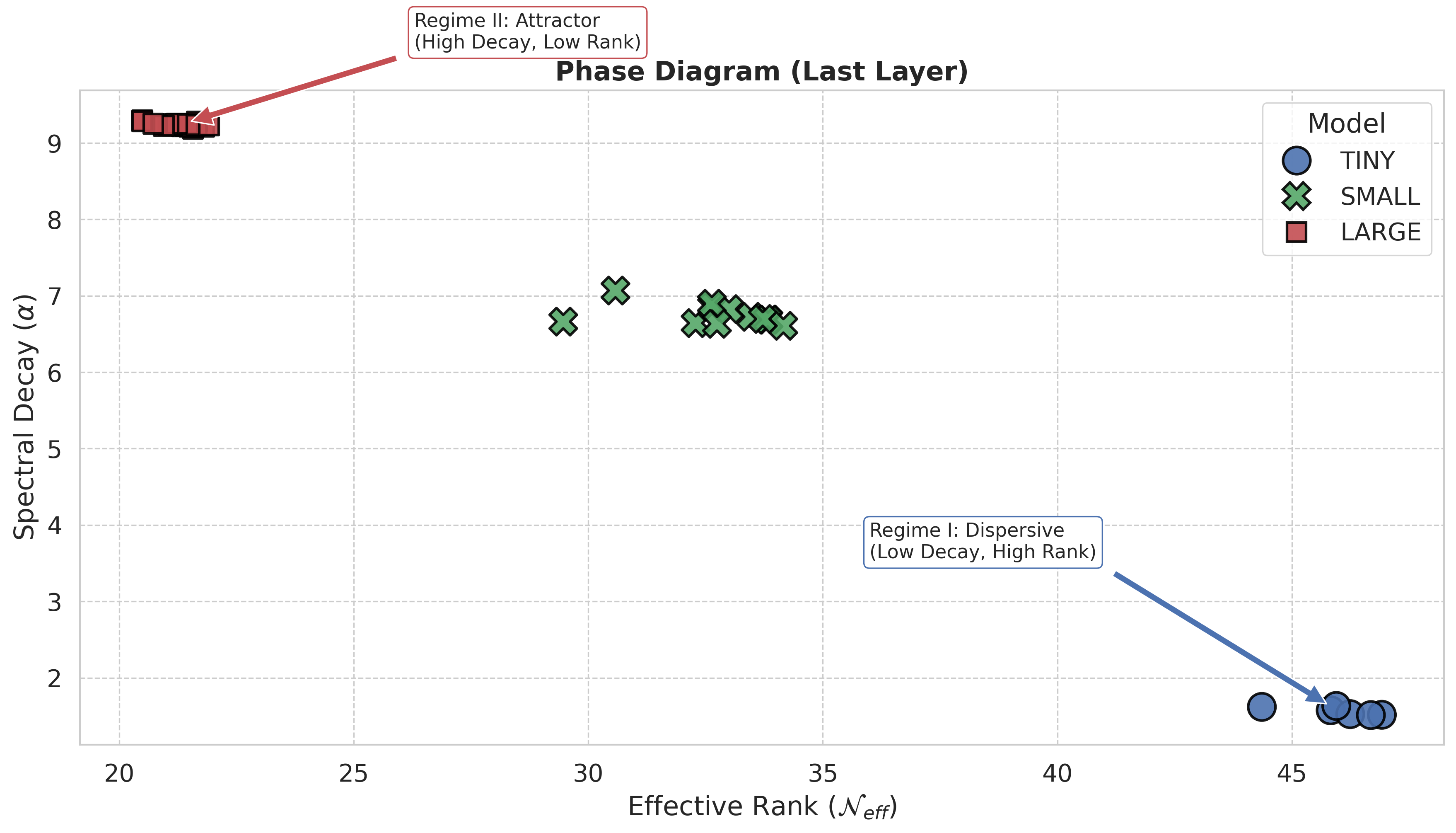}
  \caption{Spectral Phase Diagram. Scales separate monotonically from Tiny (high rank, low $\alpha$) to Large (low rank, high $\alpha$).}
  \label{fig:phase_diagram}
\end{figure}

Figure~\ref{fig:phase_diagram} presents the \textit{Spectral Phase Diagram} of the final decoder layer. We observe a fundamental separation in topology that maps directly to our theoretical regimes:
\begin{itemize}
    \item \textbf{Dispersive Cluster (Tiny/Small):} Residing in a high-rank/low-decay region, these models exhibit the spectral signatures of \textbf{Regime I}. Signal propagation is isotropic but lacks the cumulative gain to resist noise.
    \item \textbf{Attractor Cluster (Large):} Occupying a distinct low-rank/high-decay region ($\alpha > 9$), the Large model validates the conditions for \textbf{Regime II}. The high $\alpha$ (spectral decay) confirms a small spectral gap $\xi$, while low $\mathcal{N}_{\text{eff}}$ signals the onset of a rank-1 attractor state.
\end{itemize}

\subsection{Regime I: Acoustic Decoupling (Small Models)}
For intermediate scales, hallucination is driven by the collapse of the encoder-decoder interface. As shown in Figure~\ref{fig:layer_dynamics} (left), the Cross-Attention rank drops by $-13.40\%$ in Small models (Table~\ref{tab:topk50}). 
This corresponds to $\rho < 1$, where the acoustic injections $G_l$ are spectrally washed out before reaching final layers.

\subsection{Regime II: Autoregressive Lock-in (Large Models)}
\begin{figure}[t]
  \centering
  \includegraphics[width=\linewidth]{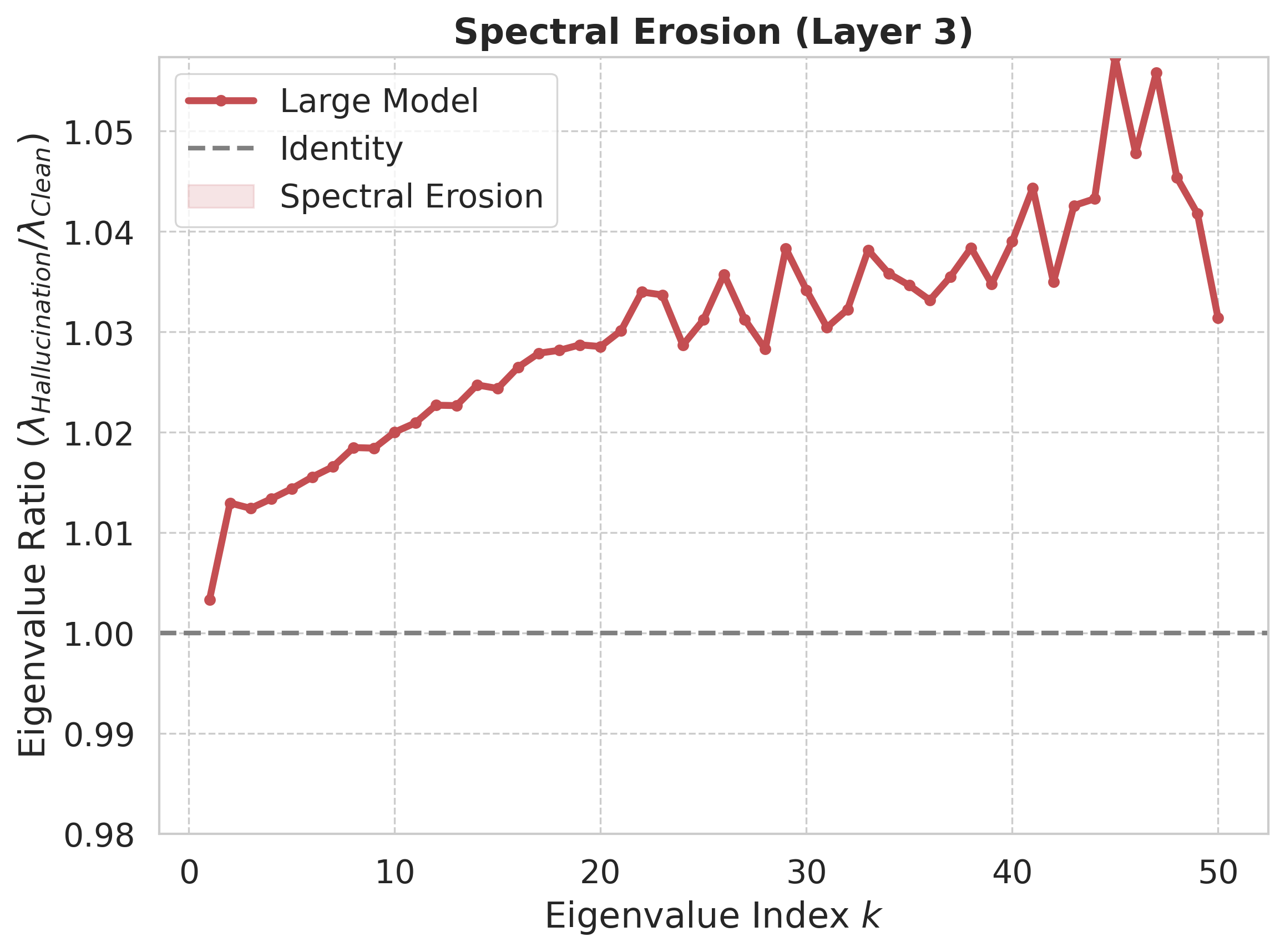}
  \caption{{Eigenvalue-ratio profile (Whisper Large, layer~3):}
$\lambda_{\text{hell}}/\lambda_{\text{clean}}$ vs.\ index $k$ --- everywhere $>1$ and
rising toward the tail (leading mode $\approx 1.003$, tail $\approx 1.04$--$1.05$).}
  \label{fig:spectral_erosion}
\end{figure}

The Large-v3-Turbo model resists global disintegration but enters an autoregressive lock-in state. The unique compression in Self-Attention rank ($-2.34\%$, Table~\ref{tab:topk50}) and the steepening of the spectral slope $\Delta \alpha > 0$ provide empirical evidence for the $u_L \Psi^T$ structure. This paradox—increased energy in tail eigenvalues (Fig.~\ref{fig:spectral_erosion}) alongside a steepening decay $\alpha$—indicates that the model actively reinforces internal hallucination modes at the expense of subtle acoustic features.

\section{Discussion}
\label{sec:discussion}

\subsection{Spectral Signatures Consistent with Theorem I.}

\textbf{Mechanism of Error:} In \textbf{Regime I}, hallucinations are symptomatic of \textit{signal dispersion}; the Jacobian norm decays ($\rho < 1$), causing the model output to decouple from acoustic evidence. In \textbf{Regime II}, hallucinations are the result of \textit{attractor capture}; the high structural alignment guides perturbed inputs into spurious basins of attraction ($\Psi$). Because the model is trapped in a rigid low-rank state, the output remains locally confident despite being acoustically unfounded. \textbf{Regime II} characterizes confidence, not correctness: distinguishing accurate from hallucinated outputs still requires task labels or external verification.

\subsection{The Alignment Assumption}
While we do not directly measure the inter-layer alignment $\kappa$ (the cosine between layer subspaces), the emergence of spectral hardening and rank compression in Large models serves as a characteristic signature of architectural alignment. Mathematically, the transition to a low-rank/high-decay regime under stress is consistent with increasing structural alignment, effectively guiding signal flow along a unified attractor manifold.

\subsection{Scale-Dependent Interpretability}
The observed phase transition challenges the intuition that hallucinations are merely ''noise.'' Our data suggests that in large models, hallucinations are \textit{over-structured}. The model becomes hypersensitive to specific internal directions ($\Psi$) while becoming blind to orthogonal evidence. 
This structural blindness reflects the inductive bias of Transformers toward low-sensitivity functions~\cite{vasudeva2025transformerslearnlowsensitivity}, where localized acoustic perturbations are suppressed in favor of global priors. As a result, standard log-probability metrics fail to detect these errors: the model is not uncertain, but geometrically constrained by its internal prior.

\section{Conclusion and Future Work}

We derived the Spectral Sensitivity Theorem and tested it on Whisper models, revealing a scale-dependent phase transition in failure modes: intermediate models exhibit \textit{Structural Disintegration} (Regime I), where acoustic signal decays in cross-attention, while large models enter \textit{Attractor Dynamics} (Regime II), marked by spectral hardening and rank compression in self-attention. Hallucinations thus reflect deterministic spectral geometry rather than stochastic noise. Our claims are empirically restricted to the Whisper family; future work will extend the analysis to other encoder-decoder ASR models such as Canary and OWSM, and exploit these signatures for real-time detection and spectral regularization against low-rank collapse.

\clearpage
\section{Generative AI Use Disclosure}
Generative AI tools were utilized solely for language refinement and editorial improvements, such as enhancing clarity, grammar, and readability. They were not used to create or alter any substantive technical material, including the conceptual framework, methodologies, experimental design, data analysis, results, figures, tables, or conclusions. All authors have reviewed and approved the final version of the manuscript and accept full responsibility for its content.
\bibliographystyle{IEEEtran}
\bibliography{mybib}

\end{document}